\documentclass{article}
\usepackage{amsmath,graphicx,mlspconf}
\usepackage{multirow}
\usepackage{algorithmic}
\usepackage{graphicx}
\usepackage{textcomp}
\usepackage{xcolor}
\usepackage{todonotes}
\usepackage{cite}
\usepackage{subcaption}
\usepackage{url}
\usepackage[symbol]{footmisc}

\toappear{2024 IEEE International Workshop on Machine Learning for Signal Processing, Sept.\ 22--25, 2024, London, UK}

\title{Improving Robustness of Spectrogram Classifiers with Neural Stochastic Differential Equations}

\name{Joel Brogan$^{\dagger}$, Olivera Kotevska$^{\dagger}$, Anibely Torres, Sumit Jha, Mark Adams}

\address{Oak Ridge National Laboratory \\
1 Bethel Valley Road, Oak Ridge, TN 37831, USA\\}

\begin{document}

\maketitle

\begin{abstract}
Signal analysis and classification is fraught with high levels of noise and perturbation. Computer-vision-based deep learning models applied to spectrograms have proven useful in the field of signal classification and detection; however, these methods aren't designed to handle the low signal-to-noise ratios inherent within non-vision signal processing tasks. While they are powerful, they are currently not the method of choice in the inherently noisy and dynamic critical infrastructure domain, such as smart-grid sensing, anomaly detection, and non-intrusive load monitoring. 
\renewcommand{\thefootnote}{\fnsymbol{footnote}} 
Currently, these models can be brittle, which makes them susceptible to noisy input. This also means they have sub-optimal stability of explanation outputs. Experts and technicians using these models to make decisions in real world scenarios need assurance that a model is performing as it is supposed to.  The classification or prediction outputs it generates should be sound and grounded, not likely to change in the presence of shifting noise landscapes. In this work, we explore the idea of Neural Stochastic Differential Equations (NSDE's) to improve the robustness of models trained to classify time series data and the effect of NSDE's on the explainability of outputs. We then test the effectiveness of these approaches by applying them to a non-intrusive load monitoring (NILM) dataset that consists of simulated harmonic signals injected into a real building. \textcolor{white}{\footnote[2]{These author Equal Author Contribution}}
\end{abstract}
\section{Introduction}


Applications of 2D deep learning methods towards efforts of signal processing and classification have been challenged by the need for more availability of sufficiently diverse data sets~\cite{padiimproved,anderson2022special}. This results in models, such as Convolutional Neural Networks (CNNs), overfitting to extraneous features of the test environment not relevant to the task at hand, learning to make "correct" classifications for the wrong reasons. The design of automated AI-based data-driven pipelines for detecting nuanced signal types and characteristics would greatly benefit from the development of algorithms to measure the confidence of neural networks in their responses. Such confidence metrics will enable human Subject Matter Experts to build a relationship of trust with robust neural networks that have a history of credible and correctly calibrated responses. In many application domains, such as load characterization, current deep learning techniques do not provide this capability in any meaningful manner \cite{alexander2020workshop,anderson2023testing}. Furthermore, many types of signal processing domains,such as jamming detection \cite{10.1109/access.2022.3150020}, speech emotion recognition \cite{10.1109/access.2022.3163712}, and radar-based classification \cite{park2020radar}, are plagued with high levels of real-world noise, either background or adversarial. We must develop techniques to combat scenarios with low Signal-to-noise (SNR) ratios. 

\begin{figure}[t]
\centering

\includegraphics[width=0.45\textwidth]{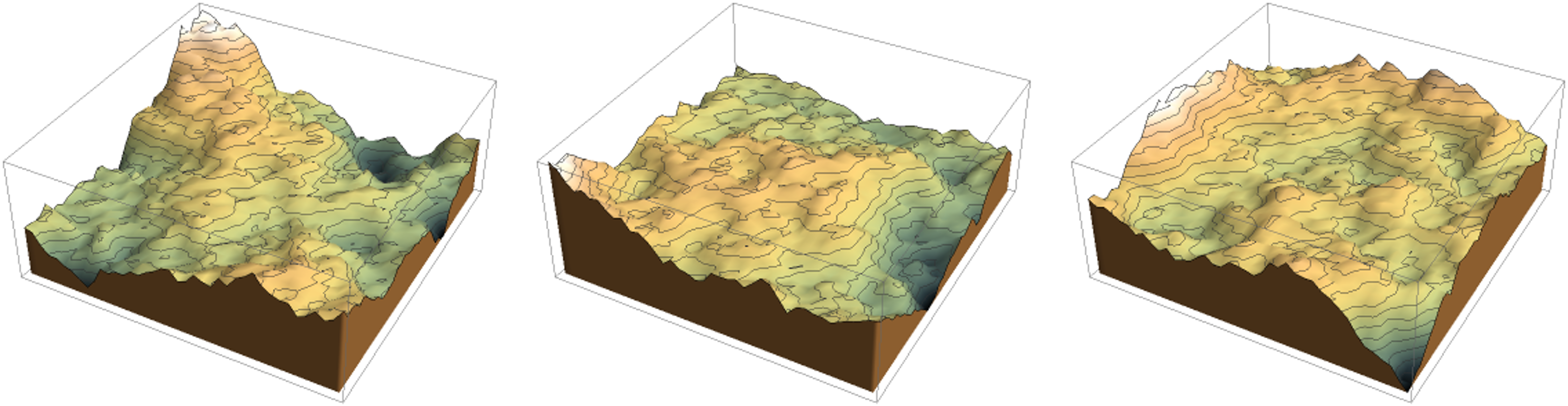}
\caption{Surface representations of the 2D Brownian surface noise injected into our Neural SDE}
\label{fig:brownian}
\vspace{-10px}
\end{figure}

This work aims to provide model training and inference methods that improve robustness to noisy spectrogram inputs, and bolster stability and confidence of subsequent classification explanation maps. The general goal is to train more robust and readily explainable neural networks by injecting appropriately shaped noise during their training. If the deep network is explainable, such that a human can get an understanding of the key features of the data that help classify an object, then SMEs can use this information to periodically verify that the network is reasoning soundly and not focusing on features specific to where the training data was collected. We formulate this noise-aware training as a Neural Stochastic Differential Equation (Neural SDE) that provides useful mathematical properties when operating in noisy domains. We test our methods on a custom-built dataset of electromagnetic waveforms injected into a building's wiring and re-collected from multiple sensors.
In this preliminary work, we contribute the following:
\begin{itemize}
    \item A new methodology for training spectrogram-domain CNNs to be robust and stable using domain-shaped noise as implemented in \cite{jha2022shaping}
    \item Preliminary experiments on Convnext and Resnet model efficacy against a non-intrusive load monitoring (NILM) dataset made from injecting signal waveforms into a building's electrical infrastructure
    \item Preliminary experiments on model efficacy against adversarial signal perturbations
    
\end{itemize}
Our results show that while modern vision-based models (specifically the ConvNeXt-Base model~\cite{woo2023convnext}) can perform well on time-series classification, they are highly susceptible to increasing levels of noise and small perturbations, while our NSDE-Convnext-variant provides competitive classification performance while incurring less performance drop as the noise floor increases. 

Our work builds on prior work on the dynamical systems models of DNNs, such as neural ODEs and neural SDEs, and stocastic NSDEs which have been investigated over the last few years \cite{chen2015learning,weinan2017proposal,lu2018beyond, wang2019resnets}. 

Chapter \ref{sec:related_work} presents the related work. Chapter \ref{sec:methods} explains the models and methods we used for the experiments and analysis. Chapter \ref{sec:experiments} describes the dataset used for experiments, processing workflow, and results. Chapter \ref{sec:conclusion} concludes the findings and presents future work.




\section{Related work}

We organized the related work into three parts:   Explainability for smart grid and spectrogram images (Sec.~\ref{sec:related_explainability}), SDEs and NSDEs (Sec.~\ref{sec:related_sdes}, and noise shaping (Sec.~\ref{sec:related_shaping}).

\subsection{Model Explainability}
\label{sec:related_explainability}
\begin{figure}[ht]
\centering

\includegraphics[width=0.4\textwidth]{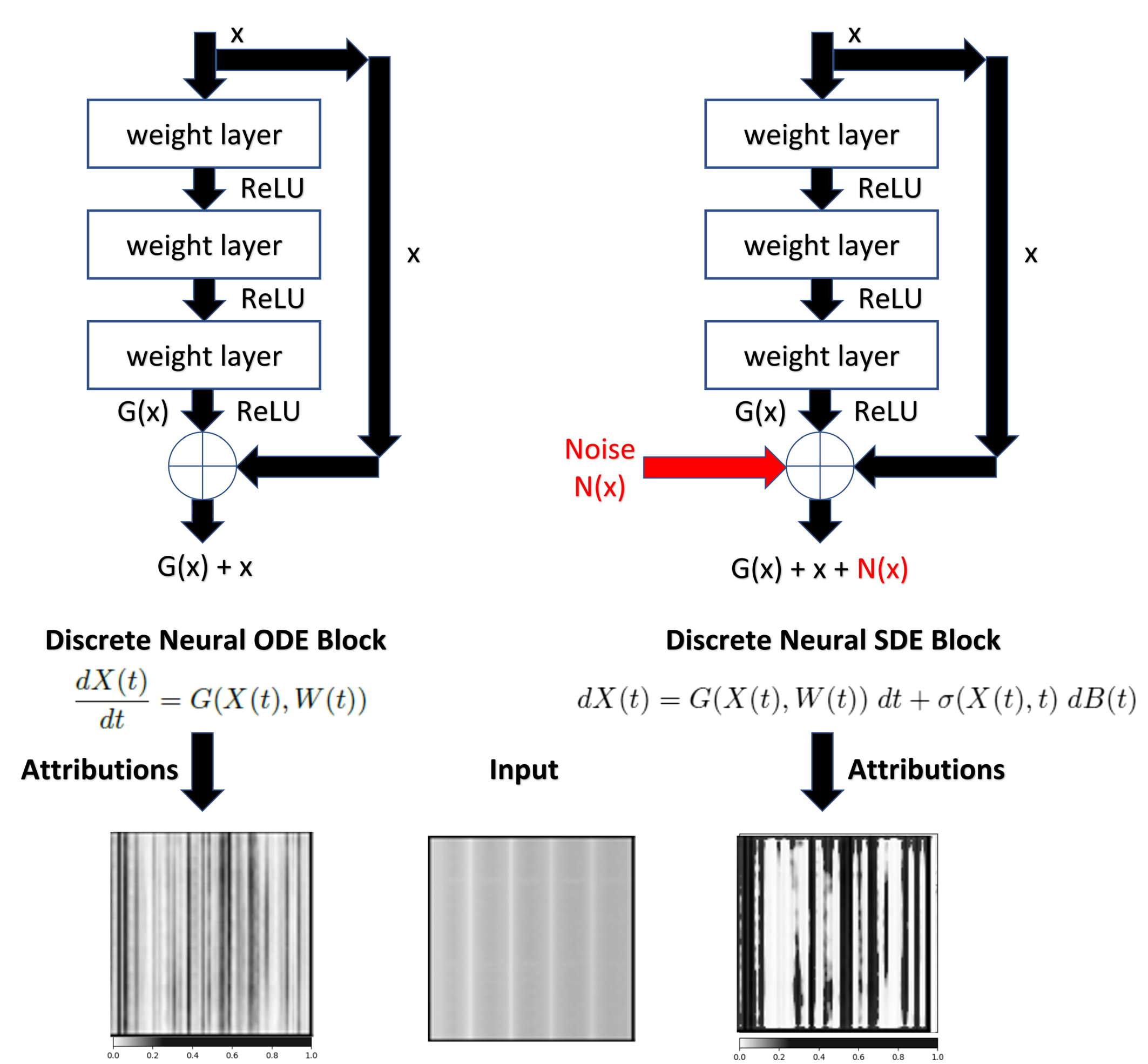}
\caption{A general overview of how shaped stochastic noise is utilized in our ConvNext architecture to produce more robust explanation attributions. }
\label{fig:workflow}
\end{figure}
In~\cite{grimes2020explanation}, the authors used the LIME (local interpretable model-agnostic explanations) tool in their classifier and succeeded in determining the frequency bands used by the classifier to make decisions about unintended radiated emission during electronic devices. Others \cite{hanchate2023explainable} used LIME to identify the critical time-frequency bands influencing the prediction of average surface roughness in a smart grinding process. LIME was used to understand the model that controlled the HVAC system \cite{kotevska2020methodology}. However, this work did not use spectrograms as input data. Some authors \cite{ferreira2021explainable} used spectrograms to analyze the magnetohydrodynamic behavior of fusion plasmas and CAM (class activation mapping) explainability tool.

Similarly, in the smart grid domain, there is a work where explainability tools were used to understand better how the model makes the decision using spectrograms as input data and what the key factors that impact those decisions \cite{xu2022review}. Others \cite{nazarytrustworthy} used Grad-CAM (gradient-weighted class activation mapping) and partial dependence plot to understand the feature's impact on fault zone prediction in smart grids. Similarly, \cite{ardito2021iscada} developed a method based on Grad-CAM that can highlight the critical regions in the spectrogram that can explain the fault type and location of the smart electric grid and provide a textual description for the event. SHAP (Shapley additive explanations) technique was used to explain the identification of faults in grid-connected photovoltaics \cite{wali2022explainable}.

\subsection{SDEs and Neural SDEs}
\label{sec:related_sdes}

A notable study by et al. \cite{10.48550} introduced Neural SDE networks, which incorporate random noise injection for regularization, enhancing the stability of Neural Ordinary Differential Equation (ODE) networks. Moreover, \cite{10.24963/ijcai.2021/73} explored smoother attributions using Neural SDEs, emphasizing reduced noise, sharper visual outcomes, and enhanced robustness of attributions computed through these models. This study highlights the benefits of employing Neural SDEs for improved interpretability and reliability in neural network applications.

\subsection{Noise Shaping}
\label{sec:related_shaping}
The shape of injected noise in an NSDE is an important factor to consider.  In \cite{kilpatrick2013wandering}, authors investigated the impact of noise on stationary pulse solutions in spatially extended neural fields, emphasizing the importance of noise shaping in neural field models.
The authors of \cite{jha2022shaping} show that particular types of noise, such as Brownian motion, result in smoother attributions and more stable explanations than traditional resnets.  \label{sec:related_work}
\vspace{-10px}
\section{Methods}
This section outlines the methods and implementations with which we obtain our results.  Generally, we implement our NSDE as a ConvNeXT architecture~\cite{woo2023convnext} and a ResNet ~\cite{he2016deep} architectures. Because these architectures contain a skip connection, they can be modeled as a Neural Differential Equation~\cite{10.48550}.  To turn a Neural DE into a Neural SDE, we must simply add shaped noise to the concatination layer of the residual block and its skip connection. Figure \ref{fig:workflow} shows how a stochastic variant of a ResNet model compares to its non-stochastic equivalent.
\vspace{-10px}
\subsection{Model Types}
Our experiments utilize a single main model - the ConvNeXt model~\cite{woo2023convnext}, which is itself a variant of the ResNet50 model.  Below we provide a short description of these model architectures and their relevance.
\textit{Residual Networks (ResNet)} were created to solve the challenge of exploiting gradients. So, the skip connections technique was developed that connects activations of a  layer to further layers by skipping some layers in between. This forms a residual block. Resnets are made by stacking these residual blocks together. The approach behind this network is that instead of layers learning the underlying mapping, we allow the network to fit the residual mapping. So, instead of say $H(x)$, initial mapping, let the network fit, $F(x) := H(x) - x$ which gives $H(x) := F(x) + x$. The advantage of adding this type of skip connection is that if any layer hurts the performance of architecture, then it will be skipped by regularization.

\textit{A ResNet-50 model}, is a 50-layer Convolutional Neural Network (CNN). The difference between ResNet50 and the previously mentioned ResNet (ResNet34) is that the building block was modified into a bottleneck design due to concerns over the time to train the layers. This used a stack of 3 layers instead of the earlier 2. Therefore, each of the 2-layer blocks in Resnet34 was replaced with a 3-layer bottleneck block, forming the Resnet 50 architecture. This results in much higher accuracy than the 34-layer ResNet model. 

\textit{ConvNeXt} is improved version of ResNet50 model. At first, a visual transformer was integrated into the model, adjusted the number of blocks at each stage, and increased the kernel size so that the sliding window did not overlap. Other changes are in the activation function, normalization task, and fewer normalization layers. ConvNeXts are good for solving general-purpose computer vision tasks, i.e., image segmentation and object detection.
\subsection{Noise Shaping}

We perform noise shaping, generation, and injection as per the implementation outlined in~\cite{jha2022shaping}. Figure~\ref{fig:brownian} shows examples of the type of brownian-shaped noise used as injection input for our Neural SDE training approach.

\subsection{Explanation Generation}
For the explainability analysis, the Captum library was used \cite{kokhlikyan2020captum}. It can be applied to any neural network model. Captum supports three types of attributions: primary, layer, and neuron. Primary attribution evaluates each input feature's contribution to a model's output. Layer attribution evaluates how a particular layer impacts the output of the model. Neuron attribution evaluates each input feature's contribution to activating a particular hidden neuron. Neuron attribution is excellent when combined with layer attribution methods because it can first inspect all the neurons in the layer. Also, a neuron attribution technique can be used to understand what a particular neuron is doing. Each category has a set of functions that provide inside information for the impact of the feature, layer, and neurons on the output. We selected Integrated Gradients (IG) \cite{sundararajan2017axiomatic} and NoiseTunnel \cite{smilkov2017smoothgrad} attributions to understand the contribution of each input feature better. 

\textit{Integrated Gradients (IG)} is a technique that aims to explain the relationship between model predictions in terms of their features by highlighting them. This is done by computing the gradients of the model's prediction output to its input features (see Equation \ref{eq:ig}). There is no need for any modification to the original deep neural network, and it can be applied to images, text, or structured data. 
In Equation \ref{eq:ig} $m$ defines a number of interpolation steps, $\partial$ is a variance of the current image, $F$ is a function representing our model, $x$ is an input, and $x^\prime$ is the baseline.

\begin{equation}
{IG}_i^{apx}(x)=(x_i-{\ x}_i^\prime)\sum_{k=1}^{m}{\frac{\partial F(x^\prime+\frac{k}{m}\times(x-x^\prime))}{\partial x_i}\ \times\frac{1}{m}}
\label{eq:ig}
\end{equation}

\textit{NoiseTunnel} is a technique that improves the accuracy of attribution methods. It adds Gaussian noise $N(0, 0.01^2)$ to each input and applies the given attribution algorithm to each sample. 

\begin{equation}
    \hat{M_{c}}(x) = \frac{1}{n}\sum_{1}^{n}M_{c}(x+N(0,\sigma ^{2}))
\label{eq:nt}
\end{equation}

\textit{Saliency maps} is another technique that highlight the most relevant regions or features within an image in a given model. 
It computes the gradient of the model's output score with respect to the input features.. The magnitude of these gradients indicates how much the output would change if the input features were modified slightly. Common methods are absolute gradient values, gradient normalization, and relevance masking. In our case absolute gradient technique was used.

\textit{Attribution-based Confidence (ABC)} metric \cite{jha2019attribution} serves as a quantitative measure to assess the reliability of deep neural network (DNN) outputs on input data. This innovative method introduces an approach to estimate DNN prediction confidence utilizing attribution techniques, such as IG, to ascertain the confidence level associated with DNN predictions. Here's how these components work together in detail Given an input sample $x$ to the DNN, IG is applied to compute the attribution map $A(x)$. This attribution map indicates the importance of each input feature in influencing the DNN's prediction for the input $x$. IG achieves this by calculating the integral of gradients of the model's output with respect to its input along a straight path from a baseline to the input of interest, providing a comprehensive understanding of feature importance across the input space. Subsequently, the ABC metric calculates a confidence score $C(x)$ based on the attribution map $A(x)$ and the predicted class label $f(x)$. The confidence score $C(x)$ is the ratio of the sum of attribution scores associated with features relevant to the predicted class to the sum of all attribution scores across all features. 

Mathematically, this can be represented as: Given an input $x$ for a model $F$ where $F_i$ denotes the $i$-th logit output of the model, we can compute attribution of feature $x_j$ of $x$ for label $i$ as $A_{j}^{i}$. We can compute ABC metric in two steps: 1) Select feature $x$ with probability $\frac{|A_{j}^{i} (x) / x_{j}|}{\sum _{j} | A_{j}^{i} (x) / x_{j}|}$ and flip the label away from $i$, that is, change the decision of the model; 2) Calculate the proportion of samples within the neighborhood where the model's decision remains consistent with the original prediction. This serves as a conservatively estimated confidence measure.

A higher value of $C(x)$ indicates stronger agreement between the predicted class label and the salient features identified by Integrated Gradients, suggesting higher confidence in the prediction. This confidence score provides valuable insights into the reliability of the DNN's predictions, enabling better decision-making based on the model's outputs.

\label{sec:methods}

\section{Experiments}\label{sec:experiments}
In this section, we will outline to dataset collected to train, test, and validate our Nueral-SDE convnext model, along with a set of experiments performed to compare model robustness to noise between the the our Neural-SDE and the original Non-Neural SDE variants of the ConvNext model. We perform three types of evaluation to show the efficacy of our Nueral-SDE method: Accuracy comparison in the presence of random noise, a robustness measure in the presence of adversarial noise, and an overall Attribute Based Confidence (ABC) comparison between the two methodologies.

\subsection{Data Source}
\label{subsec:datasource}
\begin{figure}[ht]
\centering

\includegraphics[width=0.45\textwidth]{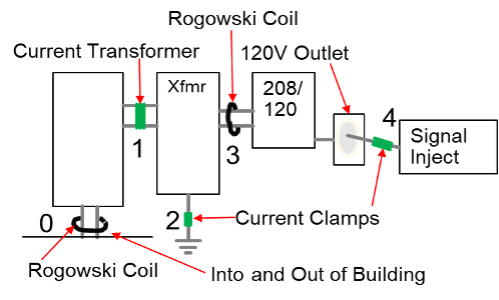}
\caption{Data collection locations in power system for injection dataset~\cite{adams2023harmonic}}
\label{fig:data_collection}
\vspace{-29px}
\end{figure}

The data, the harmonic signals dataset~\cite{adams2023harmonic} contains known signals with clean collection at injection and varying noise at other collection locations. Simulated waveforms were injected into ORNL buildings of a user facility ( Figure \ref{fig:data_collection}). The waveforms include sine, square, square (75/25 duty), and triangle waves and are injected at different frequencies (0.5-50 kHz). The signals are then subsequently measured with sensors at six locations through building power~\ref{fig:data_collection}. 16 total classes were collected, including muxed combinations of the pure waveforms, and the Short-Term Fourier transform (STFT) was used to transform the time-series collected data into a dataset of 10,000 spectrogram samples. This provides ~625 examples per class, which mimics the low-data-availability of many signal classification problems.

\subsection{Processing and Training}
The goal is to train more robust and readily explainable neural networks by injecting appropriately shaped noise during their training. The baseline models used in this work are ResNet, ResNet-50, and ConvNeXt-Base. Figure \ref{fig:workflow} presents the process overview, denoting how and where noise is injected during training to produce a Nueral-SDE from a generic res-net or ConvNext architecture. From our experimental results, we posit that ResNets with stochastic noise injected into the residual layers of our neural SDEs create more robust attributions. We show that the logarithm of the sum of the change in attributions is smaller for neural SDEs than for neural ODEs. As seen in figure~\ref{fig:workflow} and ~\ref{fig:comparison} the integrated gradient attribution with noise tunnel for our neural SDE approach (bottom right) is visually sharper than the IG attribution as well as IG coupled to a noise tunnel for the neural ODE (bottom left) The Data from~\ref{subsec:datasource} randomly partitioned into 70\%-15\%-15\% training, testing, and validation subsets, respectively.

\subsection{Results}
\textit{Experiment One} evaluates model performance in terms of accuracy in the presence of random noise \textit{and} adversarial noise robustness. We provide accuracy comparisons in Table \ref{tab:accuracy} for the baseline ConvNeXt-Base model and our Neural SDE variant. Note that the accuracy of the Neural SDE model is slightly less than that of the ConvNeXt-Base model. This is due to the regularization incurred by the noise injection, and is an expected tradeoff for robustness.

\begin{table}[ht!]
\centering
    \begin{tabular}{l|llll}
    \cline{1-3}
    \multirow{2}{*}{Model} & \multicolumn{2}{c}{Accuracy}    &  &  \\ \cline{2-3}
                  & \multicolumn{1}{l|}{Validation} & Test &  &  \\ \cline{1-3}
                  ConvNeXt-Base & \multicolumn{1}{l|}{86.9\%} & 87.20\% &  &  \\ \cline{1-3}
                  Neural SDE (ours) & \multicolumn{1}{l|}{83.2\%} & 82.88\% &  &  \\ \cline{1-3}
    \end{tabular}
    \caption{Model Accuracy}
    \label{tab:accuracy}
\end{table}

Table \ref{tab:robust_accuracy} compares how well each model holds up to Gaussian noise. We injected shaped noise from intensities of 0.05 to 1.0. The Neural SDE model's higher accuracy rates suggest greater robustness to noisy environments than the baseline model.

\begin{table}[ht!]
\centering
    \begin{tabular}{l|l|l}
    \hline
     \multirow{2}{*}{Noise} & \multicolumn{2}{c}{Accuracy} \\ \cline{2-3} 
     &           ConvNeXt-Base &  Neural SDE (ours)  \\ \hline
    0.05 &    84.38\%       &     87.50\%     \\ \hline
    0.1 &     71.88\%       &   75.00\%       \\ \hline
    0.15 &    56.25\%       &  62.50\%        \\ \hline
    0.2 &     56.25\%       &   53.12\%       \\ \hline
    0.25 &    25.00\%       &  40.62\%        \\ \hline
    0.5 &     12.50\%       &   21.88\%       \\ \hline
    0.75 &     3.12\%       &   9.38\%       \\ \hline
    1.0 &      6.25\%       &     6.25\%     \\ \hline
    \end{tabular}
    \caption{Robust Accuracy to Random Noise}
    \label{tab:robust_accuracy}
\end{table}

\textit{Experiment Two} compares model robustness to simulated adversarial noise injection attacks by using the APGD-CE (Auto-Projected Gradient Descent-Cross Entropy). We vary levels of L2 the L2 parameter on within APGD-CE and show results in Table \ref{tab:adversarial}. As can be seen, the baseline model has no robustness to this attack, while our Neural SDE still maintains some semblance of non-random classification power.

\begin{table}[ht!]
\centering
    \begin{tabular}{l|l|l}
    \hline
     \multirow{2}{*}{L2 Norm} & \multicolumn{2}{c}{Accuracy} \\ \cline{2-3} 
     &           ConvNeXt-Base &  Neural SDE (ours)  \\ \hline
    0.05 &    0\%       &     21.88\%     \\ \hline
    0.1 &     0\%       &   3.12\%       \\ \hline
    0.15 &    0\%       &  0\%        \\ \hline
    0.2 &     0\%       &   0\%       \\ \hline
    \end{tabular}
    \caption{Model Robustness to Adversarial Noise (APGD-CE)}
    \label{tab:adversarial}
\end{table}


\begin{figure}[ht]
\begin{subfigure}{.5\textwidth}
  \centering
  \includegraphics[width=.8\linewidth]{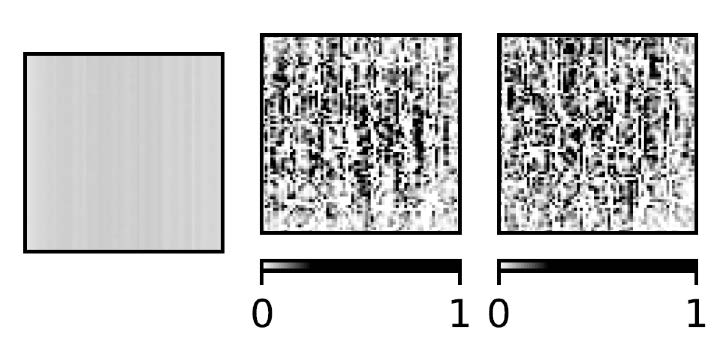}  
  \caption{Non-SDE ConvNext}
  \label{fig:sub-first}
\end{subfigure}
\begin{subfigure}{.5\textwidth}
  \centering
  \includegraphics[width=.8\linewidth]{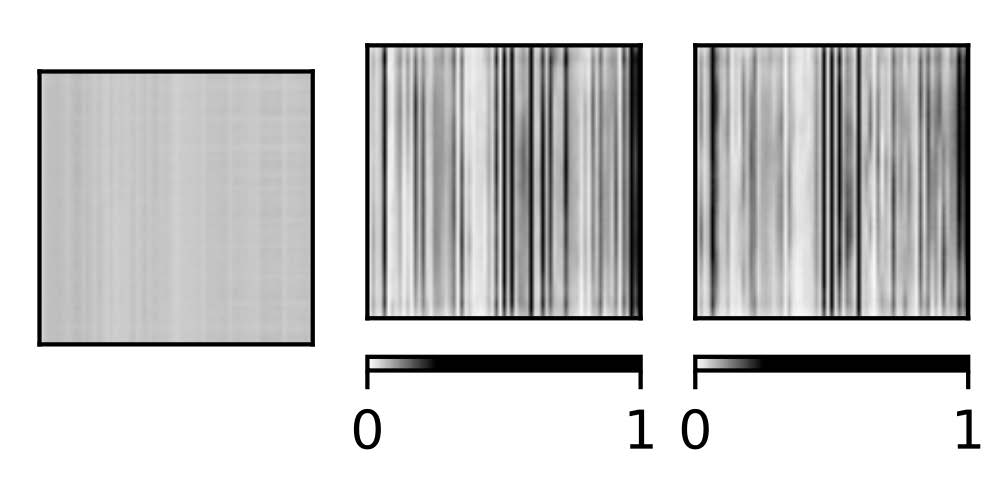}  
  \caption{Neural-SDE ConvNext}
  \label{fig:sub-second}
\end{subfigure}
\caption{A comparison of Non-Stochastic (top) and Stochastic (bottom) variants of ConvNext spectrogram explanations. A spectorgram input (left) into the Stochastic Neural SDE variant provides more coherent frequency attribution bands for both IG (center) and NoiseTunnel (right) explanations}
\label{fig:comparison}
\vspace{-20px}
\end{figure}




\textit{Experiment Three} evaluates model confidence using the  the ABC metric.  Results are presented in Table \ref{tab:abc_metric} for ConvNeXt-Base and Neural SDE models. We noticed from the results that ABC is higher for predicting the correct results for Neural SDE model then ConvNeXt, which implies higher confidence. 


\begin{table}[ht!]
\centering
    \begin{tabular}{l|l|l|l}
    \hline
    \multicolumn{2}{l|}{ConvNeXt-Base} & \multicolumn{2}{l}{Neural SDE (ours)} \\ \hline
           Correct   &  Incorrect  &   Correct  &  Incorrect      \\ \hline
            0.497    &   0.188     &   0.599    &  0.189 
    \end{tabular}
    \caption{ABC metric for correct and incorrect results}
    \label{tab:abc_metric}
\end{table}

We also evaluate explainability qualitatively, using the IG and NoiseTunnel attributions as implemented in the Captum library. A comparison of results for the 2D grayscale spectrograms explanations can be seen in Figure ~\ref{fig:comparison}. Here, we examine explainability for the 2D models trained, validated, and tested on the complete set of 2D spectrograms. In Figures \ref{fig:sub-first} and \ref{fig:sub-second}, the further left image is the original image fed into the model, the one next to it is the output explanation from IG, and  the right-most image is the explanation from noise tunnel. We can clearly notice from the results that the three methods were able to identify important virtical signal components presented in the original image when applied to the output of our Neural-SDE ConvNext variant, while the baseline contains noisey attributions.



\section{Conclusion}
From our experiments, we show that a Neural-SDE variant of the ConvNext Res-Net architecture can provide improved robustness and attribution stability for spectrogram classification in the presence of signal noise, with minimal amounts of trade-off in accuracy. This relatively minor modification to the res-net architecture can provide benifits even in low-data-availability scenarios, as shown by using our relatively small dataset for training and testing. Additionally, with Neural-SDEs, IG and Noise tunneling can successfully identify the important signal features in STFT spectrogram images. Although these preliminary results are promising, accuracy rates in the presence of noise and adversarial input are still too low to be reliable. We hope to improve the efficacy of the NSDE method by investigating improved noise shaping and training augmentation schemes that could further bolster performace in low-data-availability or few-shot-learning settings.
\label{sec:conclusion}

\section*{Acknowledgement}
\footnotesize
Notice:  This manuscript has been authored by UT-Battelle, LLC, under contract DE-AC05-00OR22725 with the US Department of Energy (DOE). The US government retains and the publisher, by accepting the article for publication, acknowledges that the US government retains a nonexclusive, paid-up, irrevocable, worldwide license to publish or reproduce the published form of this manuscript, or allow others to do so, for US government purposes. DOE will provide public access to these results of federally sponsored research in accordance with the DOE Public Access Plan
(\url{http://energy.gov/downloads/doe-public-access-plan}).

\footnotesize
\bibliographystyle{IEEEbib}
\bibliography{strings,refs}

\begin{thebibliography}{10}

\bibitem{padiimproved}
Sarala Padi, Seyed~Omid Sadjadi, Ram~D Sriram, and Dinesh Manocha,
\newblock ``Improved speech emotion recognition using transfer learning and
  spectrogram augmentation,''
\newblock in {\em Proceedings of the 2021 international conference on
  multimodal interaction}, 2021, pp. 645--652.

\bibitem{anderson2022special}
Dylan Anderson, Scott Stewart, Mark Adams, Jack Dermigny, Nathan Martindale,
  Kasimir Gabert, Boian Alexandrov, Lakshman Prasad, Joel Brogan, Zachary
  Brown, et~al.,
\newblock ``Special session on cutting edge approaches in data analytics for
  nonproliferation.,''
\newblock Tech. {R}ep., Sandia National Lab.(SNL-NM), Albuquerque, NM (United
  States), 2022.

\bibitem{alexander2020workshop}
Francis~J Alexander, Tammie Borders, Angie Sheffield, and Marc Wonders,
\newblock ``Workshop report for next-gen ai for proliferation detection:
  Accelerating the development and use of explainability methods to design ai
  systems suitable for nonproliferation mission applications,''
\newblock Tech. {R}ep., Brookhaven National Lab.(BNL), Upton, NY (United
  States); Idaho National Lab~…, 2020.

\bibitem{anderson2023testing}
Dylan Anderson, Scott Stewart, Alexei Skurikhin, Karl Pazdernik, Joel Brogan,
  and Nathan Martindale,
\newblock ``Testing and evaluation of data analytic approaches for
  nonproliferation,''
\newblock Tech. {R}ep., Oak Ridge National Laboratory (ORNL), Oak Ridge, TN
  (United States), 2023.

\bibitem{10.1109/access.2022.3150020}
Y.~Li, Jered Pawlak, J.~Price, Khair~Al Shamaileh, Quamar Niyaz, Paheding
  Sidike, and Vijay Devabhaktuni,
\newblock ``Jamming detection and classification in ofdm-based uavs via
  feature- and spectrogram-tailored machine learning,''
\newblock {\em Ieee Access}, 2022.

\bibitem{10.1109/access.2022.3163712}
Ammar Amjad, Lal Khan, Noman Ashraf, Muhammad~Bilal Mahmood, and Hsien-Tsung
  Chang,
\newblock ``Recognizing semi-natural and spontaneous speech emotions using deep
  neural networks,''
\newblock {\em Ieee Access}, 2022.

\bibitem{park2020radar}
Dongsuk Park, Seungeui Lee, SeongUk Park, and Nojun Kwak,
\newblock ``Radar-spectrogram-based uav classification using convolutional
  neural networks,''
\newblock {\em Sensors}, vol. 21, no. 1, pp. 210, 2020.

\bibitem{jha2022shaping}
Sumit~Kumar Jha, Rickard Ewetz, Alvaro Velasquez, Arvind Ramanathan, and Susmit
  Jha,
\newblock ``Shaping noise for robust attributions in neural stochastic
  differential equations,''
\newblock in {\em Proceedings of the AAAI Conference on Artificial
  Intelligence}, 2022, vol.~36, pp. 9567--9574.

\bibitem{woo2023convnext}
Sanghyun Woo, Shoubhik Debnath, Ronghang Hu, Xinlei Chen, Zhuang Liu, In~So
  Kweon, and Saining Xie,
\newblock ``Convnext v2: Co-designing and scaling convnets with masked
  autoencoders,''
\newblock in {\em Proceedings of the IEEE/CVF Conference on Computer Vision and
  Pattern Recognition}, 2023, pp. 16133--16142.

\bibitem{chen2015learning}
Yunjin Chen, Wei Yu, and Thomas Pock,
\newblock ``On learning optimized reaction diffusion processes for effective
  image restoration,''
\newblock in {\em Proceedings of the IEEE conference on computer vision and
  pattern recognition}, 2015, pp. 5261--5269.

\bibitem{weinan2017proposal}
Ee~Weinan,
\newblock ``A proposal on machine learning via dynamical systems,''
\newblock {\em Communications in Mathematics and Statistics}, vol. 1, no. 5,
  pp. 1--11, 2017.

\bibitem{lu2018beyond}
Yiping Lu, Aoxiao Zhong, Quanzheng Li, and Bin Dong,
\newblock ``Beyond finite layer neural networks: Bridging deep architectures
  and numerical differential equations,''
\newblock in {\em International Conference on Machine Learning}. PMLR, 2018,
  pp. 3276--3285.

\bibitem{wang2019resnets}
Bao Wang, Zuoqiang Shi, and Stanley Osher,
\newblock ``Resnets ensemble via the feynman-kac formalism to improve natural
  and robust accuracies,''
\newblock {\em Advances in Neural Information Processing Systems}, vol. 32,
  2019.

\bibitem{grimes2020explanation}
Tom Grimes, Eric Church, William Pitts, and Lynn Wood,
\newblock ``Explanation of unintended radiated emission classification via
  lime,''
\newblock {\em arXiv preprint arXiv:2009.02418}, 2020.

\bibitem{hanchate2023explainable}
Abhishek Hanchate, Satish~TS Bukkapatnam, Kye~Hwan Lee, Anil Srivastava, and
  Soundar Kumara,
\newblock ``Explainable ai (xai)-driven vibration sensing scheme for surface
  quality monitoring in a smart surface grinding process,''
\newblock {\em Journal of Manufacturing Processes}, vol. 99, pp. 184--194,
  2023.

\bibitem{kotevska2020methodology}
Olivera Kotevska, Jeffrey Munk, Kuldeep Kurte, Yan Du, Kadir Amasyali, Robert~W
  Smith, and Helia Zandi,
\newblock ``Methodology for interpretable reinforcement learning model for hvac
  energy control,''
\newblock in {\em 2020 IEEE International Conference on Big Data (Big Data)}.
  IEEE, 2020, pp. 1555--1564.

\bibitem{ferreira2021explainable}
Diogo~R Ferreira, Tiago~A Martins, Paulo Rodrigues, and JET Contributors,
\newblock ``Explainable deep learning for the analysis of mhd spectrograms in
  nuclear fusion,''
\newblock {\em Machine Learning: Science and Technology}, vol. 3, no. 1, pp.
  015015, 2021.

\bibitem{xu2022review}
Chongchong Xu, Zhicheng Liao, Chaojie Li, Xiaojun Zhou, and Renyou Xie,
\newblock ``Review on interpretable machine learning in smart grid,''
\newblock {\em Energies}, vol. 15, no. 12, pp. 4427, 2022.

\bibitem{nazarytrustworthy}
Fatemeh Nazary, Carmelo Ardito, Eugenio Di~Sciascio, Eng~Gianluca Sapienza, and
  Mario Carpentieri,
\newblock ``Trustworthy machine learning in smart grids,''
\newblock .

\bibitem{ardito2021iscada}
Carmelo Ardito, Yashar Deldjoo, Eugenio Di~Sciascio, Fatemeh Nazary, and
  Gianluca Sapienza,
\newblock ``Iscada: Towards a framework for interpretable fault prediction in
  smart electrical grids,''
\newblock in {\em IFIP Conference on Human-Computer Interaction}. Springer,
  2021, pp. 270--274.

\bibitem{wali2022explainable}
Syed Wali and Irfan Khan,
\newblock ``Explainable signature-based machine learning approach for
  identification of faults in grid-connected photovoltaic systems,''
\newblock in {\em 2022 IEEE Texas Power and Energy Conference (TPEC)}. IEEE,
  2022, pp. 1--6.

\bibitem{10.48550}
X.~Liu,
\newblock ``Neural sde: stabilizing neural ode networks with stochastic
  noise,''
\newblock 2019.

\bibitem{10.24963/ijcai.2021/73}
S.~Jha, R.~Ewetz, A.~Velasquez, and S.~Jha,
\newblock ``On smoother attributions using neural stochastic differential
  equations,''
\newblock 2021.

\bibitem{kilpatrick2013wandering}
Zachary~P Kilpatrick and Bard Ermentrout,
\newblock ``Wandering bumps in stochastic neural fields,''
\newblock {\em SIAM Journal on Applied Dynamical Systems}, vol. 12, no. 1, pp.
  61--94, 2013.

\bibitem{he2016deep}
Kaiming He, Xiangyu Zhang, Shaoqing Ren, and Jian Sun,
\newblock ``Deep residual learning for image recognition,''
\newblock in {\em Proceedings of the IEEE conference on computer vision and
  pattern recognition}, 2016, pp. 770--778.

\bibitem{kokhlikyan2020captum}
Narine Kokhlikyan, Vivek Miglani, Miguel Martin, Edward Wang, Bilal Alsallakh,
  Jonathan Reynolds, Alexander Melnikov, Natalia Kliushkina, Carlos Araya, Siqi
  Yan, et~al.,
\newblock ``Captum: A unified and generic model interpretability library for
  pytorch,''
\newblock {\em arXiv preprint arXiv:2009.07896}, 2020.

\bibitem{sundararajan2017axiomatic}
Mukund Sundararajan, Ankur Taly, and Qiqi Yan,
\newblock ``Axiomatic attribution for deep networks,''
\newblock in {\em International conference on machine learning}. PMLR, 2017,
  pp. 3319--3328.

\bibitem{smilkov2017smoothgrad}
Daniel Smilkov, Nikhil Thorat, Been Kim, Fernanda Vi{\'e}gas, and Martin
  Wattenberg,
\newblock ``Smoothgrad: removing noise by adding noise,''
\newblock {\em arXiv preprint arXiv:1706.03825}, 2017.

\bibitem{jha2019attribution}
Susmit Jha, Sunny Raj, Steven Fernandes, Sumit~K Jha, Somesh Jha, Brian
  Jalaian, Gunjan Verma, and Ananthram Swami,
\newblock ``Attribution-based confidence metric for deep neural networks,''
\newblock {\em Advances in Neural Information Processing Systems}, vol. 32,
  2019.

\bibitem{adams2023harmonic}
Mark~B Adams, Gregory Sheets, Philip Bingham, Mason Taylor, Michelle Baldwin,
  and Scott~L Stewart,
\newblock ``Harmonic signals dataset,''
\newblock Tech. {R}ep., Oak Ridge National Lab.(ORNL), Oak Ridge, TN (United
  States). Oak Ridge~…, 2023.

\end{thebibliography}

\end{document}